\documentclass[letterpaper, 10 pt, conference]{ieeeconf}  

\IEEEoverridecommandlockouts                              

\usepackage{graphicx} 

\usepackage{fancyhdr}
\fancypagestyle{withfooter}{
  
  \fancyfoot[C]{\footnotesize Accepted to the IEEE ICRA Workshop on Field Robotics 2024}
}

\usepackage{amsmath} 
\usepackage{amssymb}  
\usepackage[table]{xcolor}
\usepackage{colortbl}
\usepackage{booktabs}
\usepackage{balance} 

\newcommand{\BibTeX}{\rm B\kern-.05em{\sc i\kern-.025em b}\kern-.08em\TeX}

\newcommand{\omitit}[1]{}

\title{Evaluating Collaborative Autonomy in Opposed Environments using Maritime Capture-the-Flag Competitions}

\author{Jordan Beason$^{1}$, Michael Novitzky$^{1}$, John Kliem $^{2}$, Tyler Errico$^{1}$, Zachary Serlin$^{3}$, Kevin Becker$^{4}$,\\ Tyler Paine$^{4,6}$, Michael Benjamin$^{4}$, Prithviraj Dasgupta$^{2}$, Peter Crowley$^{5}$, Charles O'Donnell$^{1}$, John James$^{1}$
\thanks{$^{1}$United States Military Academy, West Point, NY,
        Corresponding author: {\tt\small jordan.beason@westpoint.edu}}
\thanks{$^{2}$Naval Research Laboratory, Washington, D.C}%
\thanks{$^{3}$MIT Lincoln Laboratory, Lexington, MA}%
\thanks{$^{4}$Massachusetts Institute of Technology, Cambridge, MA}%
\thanks{$^{5}$Boston University, Boston, MA}%
\thanks{$^{6}$Woods Hole Oceanographic Institution, Woods Hole, MA}%
}

\begin{document}

\maketitle
\thispagestyle{withfooter}
\pagestyle{withfooter}

\begin{abstract}
The objective of this work is to evaluate multi-agent artificial intelligence methods when deployed on teams of unmanned surface vehicles (USV) in an adversarial environment. Autonomous agents were evaluated in real-world scenarios using the Aquaticus test-bed, which is a Capture-the-Flag (CTF) style competition involving teams of USV systems. Cooperative teaming algorithms of various foundations in behavior-based optimization and deep reinforcement learning (RL) were deployed on these USV systems in two versus two teams and tested against each other during a competition period in the fall of 2023. Deep reinforcement learning applied to USV agents was achieved via the Pyquaticus test bed, a lightweight gymnasium environment that allows simulated CTF training in a low-level environment. The results of the experiment demonstrate that rule-based cooperation for behavior-based agents outperformed those trained in Deep-reinforcement learning paradigms as implemented in these competitions. Further integration of the Pyquaticus gymnasium environment for RL with MOOS-IvP in terms of configuration and control schema will allow for more competitive CTF games in future studies. As the development of experimental deep RL methods continues, the authors expect that the competitive gap between behavior-based autonomy and deep RL will be reduced. As such, this report outlines the overall competition, methods, and results with an emphasis on future works such as reward shaping and sim-to-real methodologies and extending rule-based cooperation among agents to react to safety and security events in accordance with human experts intent/rules for executing safety and security processes. 
\end{abstract}




\section{Introduction}
 Robots and other unmanned vehicles are continuing to become more widespread in both civilian and military applications. In recent years, unmanned vehicles have been deployed in ground, air, surface, and underwater domains, with the term ``UxV'' serving as an encapsulating term for various autonomous agents working within these domains. Past trends in robotics and unmanned systems have mainly focused on the deployment of singular, teleoperated UxV's. With advances developing in autonomy and distributed systems, recent focus has been given to the evolution of multi-agent UxV swarms, which utilize artificial intelligence (AI) for autonomous operation as opposed to human teleoperation. As the use of unmanned swarms becomes more widespread in the civilian and defense sectors, it is vital to challenge the assumption that AI-based systems will work in an environment where other autonomous agents are benign and cooperative \cite{Gupta2018}. 

\begin{figure}[h]
  \centering
  \vspace{5pt}
  \includegraphics[width=0.8\linewidth]{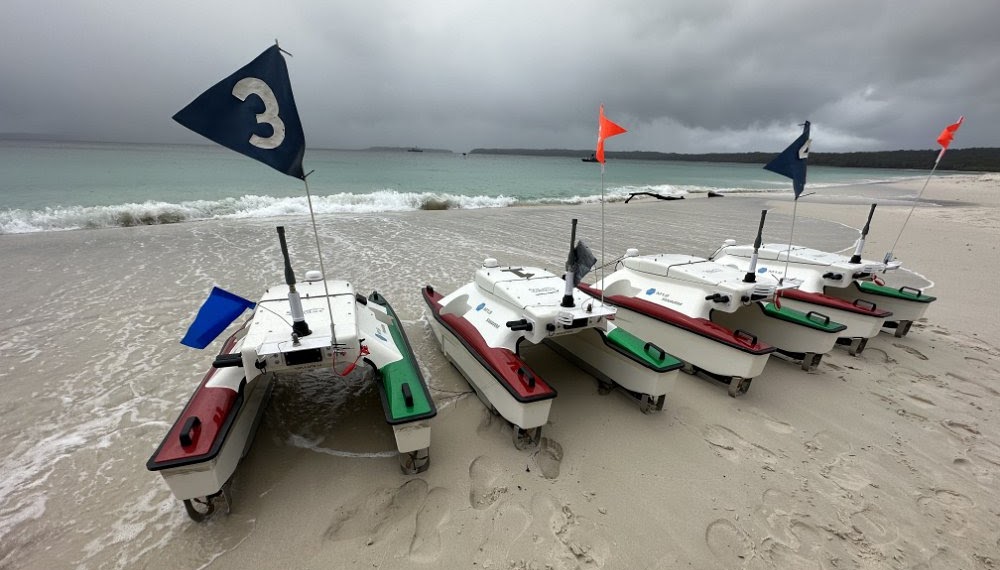} 
  \vspace{6pt}
  \caption{Sea Robotics Surveyor Class USV's from field experiments in October 2023.}
  \label{fig:USV}
\end{figure}
 The objective of this work is to evaluate multi-agent artificial intelligence methods deployed on UxV's in an adversarial environment. Autonomous agents were evaluated in real-world scenarios using the Aquaticus \cite{Novitzky2019} test-bed, which is a Capture-the-Flag (CTF) style competition involving teams of unmanned surface vehicles (USVs), shown in \ref{fig:USV}. Cooperative teaming algorithms of various foundations in hierarchical behavior trees, interval programming \cite{Newman2008}, and deep reinforcement learning (RL) \cite{Gupta2018} were deployed on these USV systems and tested against each other during a two-week competition period during the fall of 2023. This paper describes an overview of the Project Aquaticus test-bed, the methods with which various artificial intelligence methods were developed, and their respective results when utilized in an adverse environment with incomplete information of the environment state. Implications from the experiment are discussed from a mostly qualitative perspective, as future work is required to evolve these cooperative teaming methods for more competitive algorithms in future Aquaticus experiments.

\section{Motivation}
Research on cooperative teaming within the context of a USV CTF competition fills a number of novel research gaps. The primary motivation of this work is to investigate and bridge the discrepancies between simulated and field autonomy. As such, it is important to evaluate AI methods that are adaptable to non-observable aspects of their environment, such as constrained sensory perception or adversary intent. Another important catalyst for this research is the intention to adapt deep reinforcement learning (RL) technology commonly used in real-time strategy (RTS) games and compose similar methods for use within the Aquaticus test-bed ~\cite{spencer2021opposed}, \cite{Kliem2023}. RTS games have challenges similar to CTF competition, such as sparse rewards, large state-action spaces, and limited observability of the game state. Deep RL within RTS games has achieved better intelligence than human experts within a number of studies, with various levels of complexity from 1 vs. 1 checkers to 200 vs. 200 matches of popular RTS games like StarCraft-II \cite{Gupta2018}. This research explores the first initial steps in evaluating Deep RL enabled agents against traditional behavior and decision tree-based agents with disputing end goals. 

\section{The Project Aquaticus Test-bed}
The Project Aquaticus test-bed originated on the Charles River at the Massachusetts Institute of Technology in the summer of 2015. Until 2019, Project Aquaticus focused predominantly on human-robot interaction, levels of autonomy, and trust in zero-sum games such as capture the flag. In those experiments, a team would consist of two robots and two humans in motorized kayaks. In 2019, the Project Aquaticus sandbox was migrated to the United States Military Academy and incorporated into the multi-domain human-robot teaming sandbox (MDO-HuRT-S).

\subsection{Game Setup and Rules}
\begin{figure}[h]
    \centering
    \includegraphics[width=.8\linewidth]{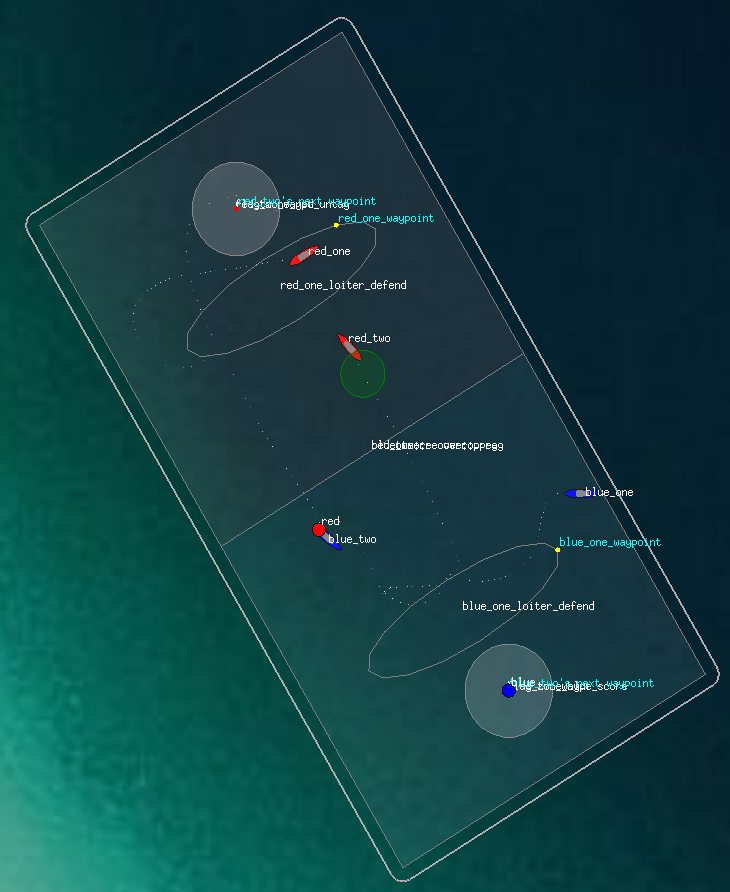}
    \includegraphics[width=0.8\linewidth]{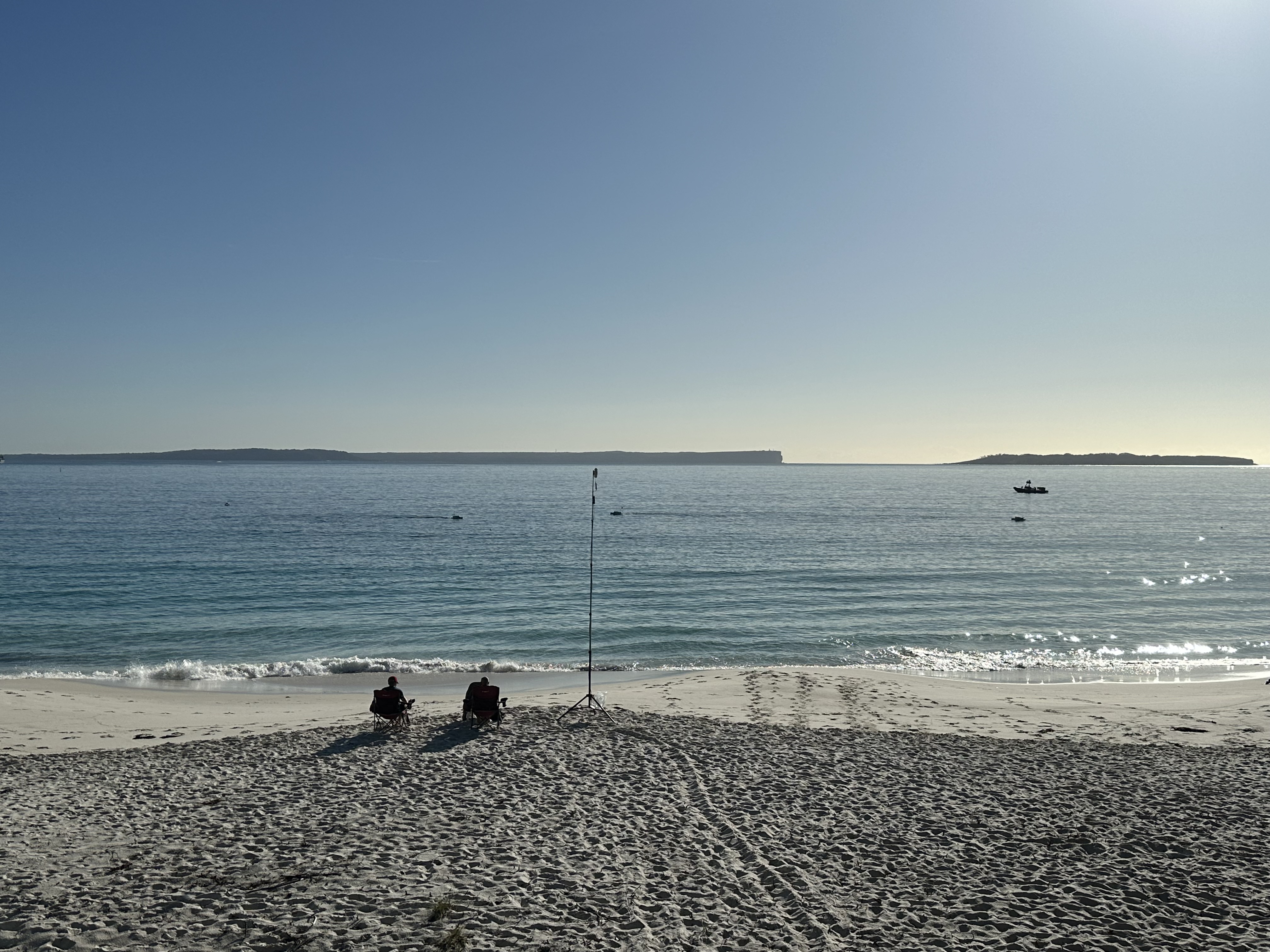}
    \caption{Above: Playback of an Aquaticus CTF game, red two is tagged and returning to base, while blue two is returning to base with a "grabbed" flag. Below: The ocean gameplay environment }
    \label{fig:EXAMPLE_AQUATICUS_GAME}
\end{figure}

Project Aquaticus utilizes capture the flag as the zero-sum game of choice. In general, the goal of an agent in Project Aquaticus is to \emph{GRAB} an opponent's flag and return it to their own home location, which is called a \emph{CAPTURE}. A grab is given +1 point, and a subsequent capture is given +2 points.  The defending agents are allowed to \emph{TAG} an opponent if they themselves are untagged and on their own side of the field. An agent who leaves the boundaries of the playing field will be automatically tagged. An agent carrying a flag that is then tagged will lose the flag as it will be returned to its own home location. In order for an agent to untag themselves, they must return to their own home-flag region. A mission replay of an Aquaticus game illustrating the game environment is depicted in Figure \ref{fig:EXAMPLE_AQUATICUS_GAME}, as well as the coastal setting for experiments conducted in October 2023.  

\subsection{An Introduction to MOOS-IvP}
MOOS-IvP includes two components: Mission Oriented Operating Suite (MOOS) middleware and Interval Programming (IvP) decision-making optimization \cite{Benjamin2010}.  MOOS is an open-source robot publish-subscribe middleware that is lightweight, only requiring the minimal dependencies. The general architecture of MOOS is a centralized database called MOOSDB within a MOOS community. In this work a MOOS community exists on each robot and the shore-side computer. MOOS applications on each robot interface with sensors, communications, and the Searobotics Surveyor USV front seat computer that controls thrusters, GPS, and compass. 
The entire MOOS-IvP ecosystem has the capability for mission planning, mission execution, and post-mission tools for autonomous robots. All figures of CTF gameplay in this report are live playbacks of the experiments using the alogview application. 

\begin{figure}[h]
    \centering
    \includegraphics[width=.7\linewidth]{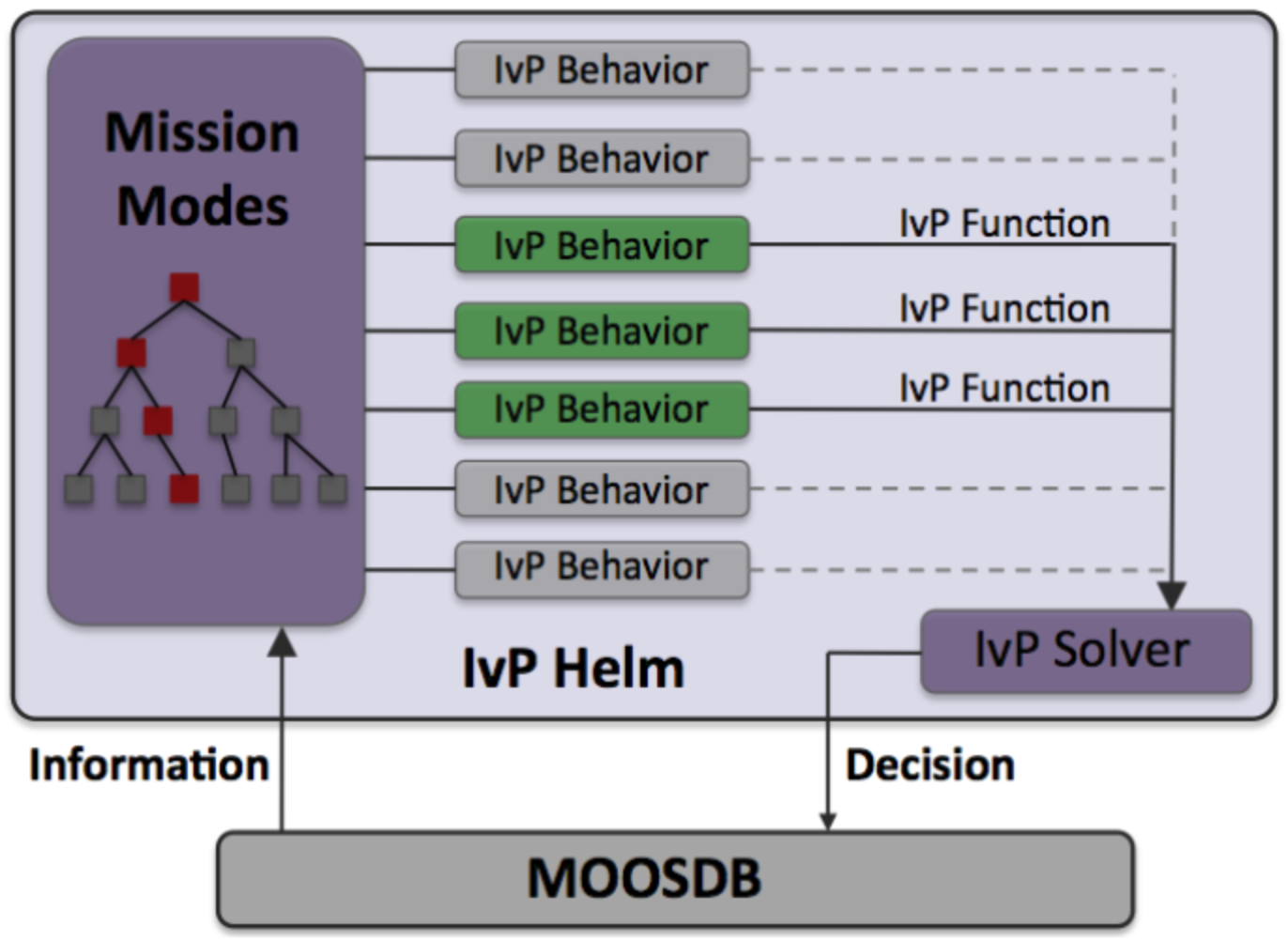}
    \caption{A visualization of the IvP Helm arcitecture.}
    \label{fig:MOOSIVP}
\end{figure}

Of particular interest in this paper is the MOOS-IvP application pHelmIvP. The pHelmIvP includes a behavior-based optimizer that can solve multiple behaviors simultaneously, such as the \emph{Waypoint}, \emph{Loiter}, and \emph{Avoid Collision} behaviors. Concurrent behaviors are grouped into a hierarchy of modes, which can then be turned on and off based on which mode is active. For example, a maritime surface robot may be making way to a point with \emph{GoToWaypoint} behavior but also have the \emph{AvoidCollision} behavior. Once the robot arrives at the point, it might switch modes to one in which only the \emph{StationKeep} behavior is active. Figure \ref{fig:MOOSIVP} shows an overview of the pHelmIvP architecture.   

Entries in the Project Aquaticus competition can leverage MOOS-IvP for its autonomy as a combination of pHelmIvP behaviors and modes along with MOOS applications for coordination among all agents on a team. All entries, regardless of using only MOOS-IvP or in combination with PyQuaticus were required to have two behaviors concurrently running at all times: \emph{OpRegion} and \emph{AvoidCollision}. This was to ensure overall game safety regardless of errors in competition entries or unforeseen interactions. The \emph{OpRegion} behavior ensures that when a USV leaves the field of the game, it will reenter the field safely and slowly. This is a layer of safety within the autonomy to ensure that USVs don't go off beyond the game field and the range of the safety remote controls if needed. The \emph{AvoidCollision} parameters were set as a last resort if an entry erroneously got too close to other USVs. In the Project Aquaticus games, the \emph{AvoidCollision} was set to halt all motion such that the safety controllers can de-conflict the agents and continue the game.

The publicly accessible version of MOOS-IvP Aquaticus incorporates a set of predefined roles, among which users can choose:
\begin{itemize}
    \item \textbf{Easy Attacker:}
    The USV directly navigates toward the flag, disregarding the positions of both opponents and teammates. Upon securing the flag or being tagged, it returns to its base.
    \item \textbf{Easy Defender:}
    The USV orbits its own flag in a protective maneuver, remaining unresponsive to external stimuli and focusing only on flag defense.
    \item \textbf{Medium Attacker:}
    This behavior involves the USV advancing towards the flag while concurrently evading opponents.
    \item \textbf{Medium Defender:}
    The USV maintains a central position within its territory until an adversary approaches. Subsequently, it engages the opponent, pursuing them until they are tagged or retreat across the field.
\end{itemize}

\begin{figure}[h]
    \centering
    \includegraphics[width=.75\linewidth]{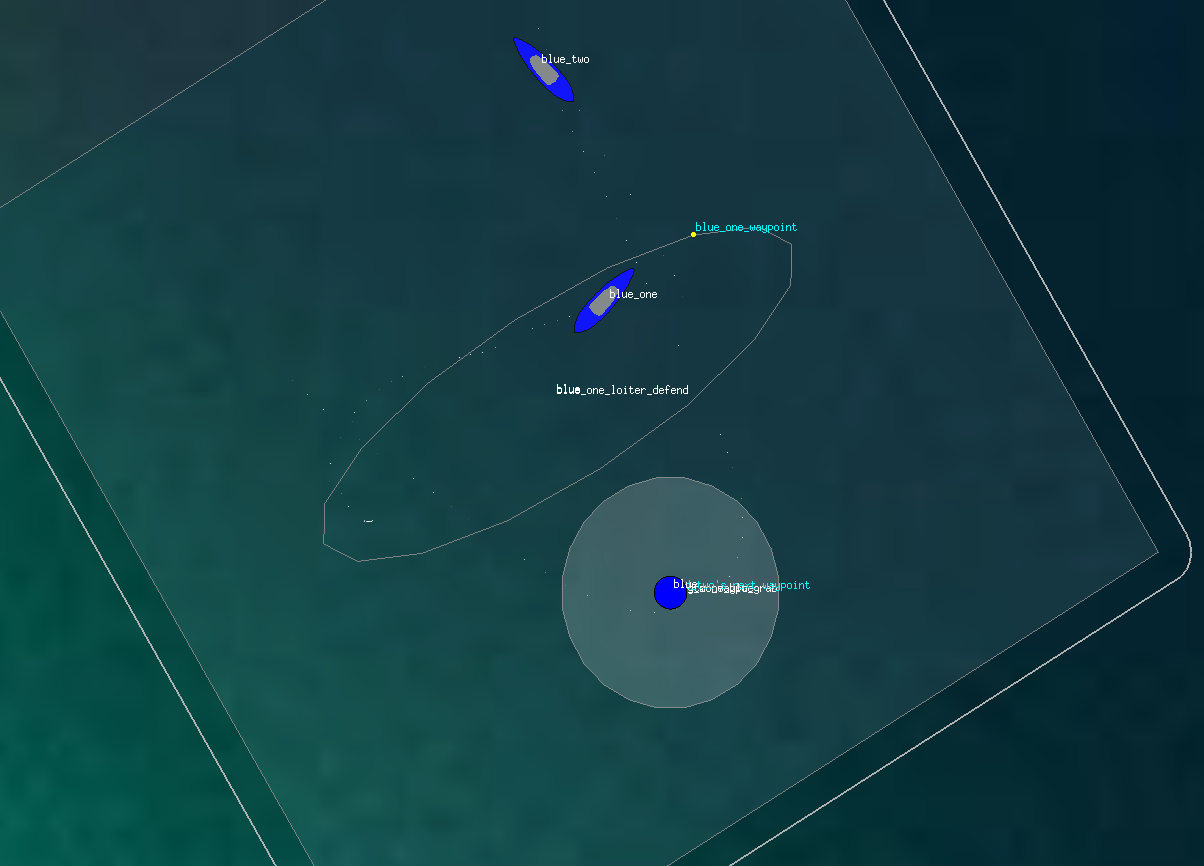}
    \caption{An Example of the default Pav01 behavior, one attacker boat and a loiter behavior defense agent.}
    \label{fig:AQUATICUS_DEFAULT_OPPONENT}
\end{figure}

\section{Entries}
This section describes the two flavors of entries. All entries leveraged MOOS-IvP as the autonomy stack running on the SeaRobotics Survyor USVs. The first set of entries leveraged MOOS-IvP behaviors that came out of the box.  The Pyquaticus entries leveraged the ability to use reinforcement learning to learn a behavior that could then be executed through MOOS-IvP.

\subsection{Behavior Based Autonomy using MOOS-IvP}

These entries used the following two features of the behavior-based autonomy framework in pHelmIvP: mode selection and multi-objective behavior optimization of multiple active behaviors.   

\subsubsection{Pav01 - The Default Aquaticus Opponent}\label{sec:Pav01}


The default baseline opponent in the Aquaticus competition is called Pav01. In the Pav01 configuration, a human randomly assigns agents as attackers or defenders before competition, and does not change roles nor react to the opponents. An agent is assigned as \textbf{Easy Attacker}, persistently attempts to capture the opposing flag without avoiding enemy agents. The other agent is assigned as the \textbf{Easy Defender}, it circles in front of the flag using a cyclic waypoint and does not go out of its way to persecute adversary agents in the area. This basic implementation, which does not take full advantage of pHelmIvP's capabilities, serves as a baseline opponent that will be used to fairly evaluate other entries of the competition. Figure \ref{fig:AQUATICUS_DEFAULT_OPPONENT} shows a mission replay of a Pav01 USV team in game. 

On each vehicle, the mode selection process selects a pre-determined role, and in this approach, the role can be generically classified as either an attacker or a defender.  
The mode selection process is a sequence of logic statements based on the current state of the game. For example, the role of defender would be to pursue an intruder if an opposing vehicle has entered the home side of the field. This role assignment is flexible and can change depending on the state of the game. 
Once a mode is selected, a predefined collection of behaviors corresponding to the role becomes active.
In this work, a combination of standard MOOS-IvP behaviors: the Waypoint Behavior, the Loiter Behavior, and the Cut Range Behavior were used in addition to the two mandatory behaviors of OpBoxRegion and Avoid Collision. Both agents were configured with the same mission file but with different triggers. In fact, no code was necessary to design these strategies because all logic was included as configuration parameters for behaviors in MOOS-IvP. 
%

These strategies require a baseline of information about the state of the game. First, it assumes both flag positions are known. The defender would also need to have the position of any not-tagged intruders to be able to pursue them. 
Finally, the attacker would need to know the position of any enemy defenders to perform evasive maneuvers.

\subsubsection{Static Roles and Rule-based Switching}
The remainder of this subsection describes a progression of strategies (2-4) that built upon the (very) basic default strategy, Pav01, described in Section \ref{sec:Pav01}. These iterations were completed in the first few days of the real-world competition, and the final version, strategy 4, remained unchanged through the end of the competition.  

\begin{figure}[h]
    \centering
    \includegraphics[width=1\linewidth]{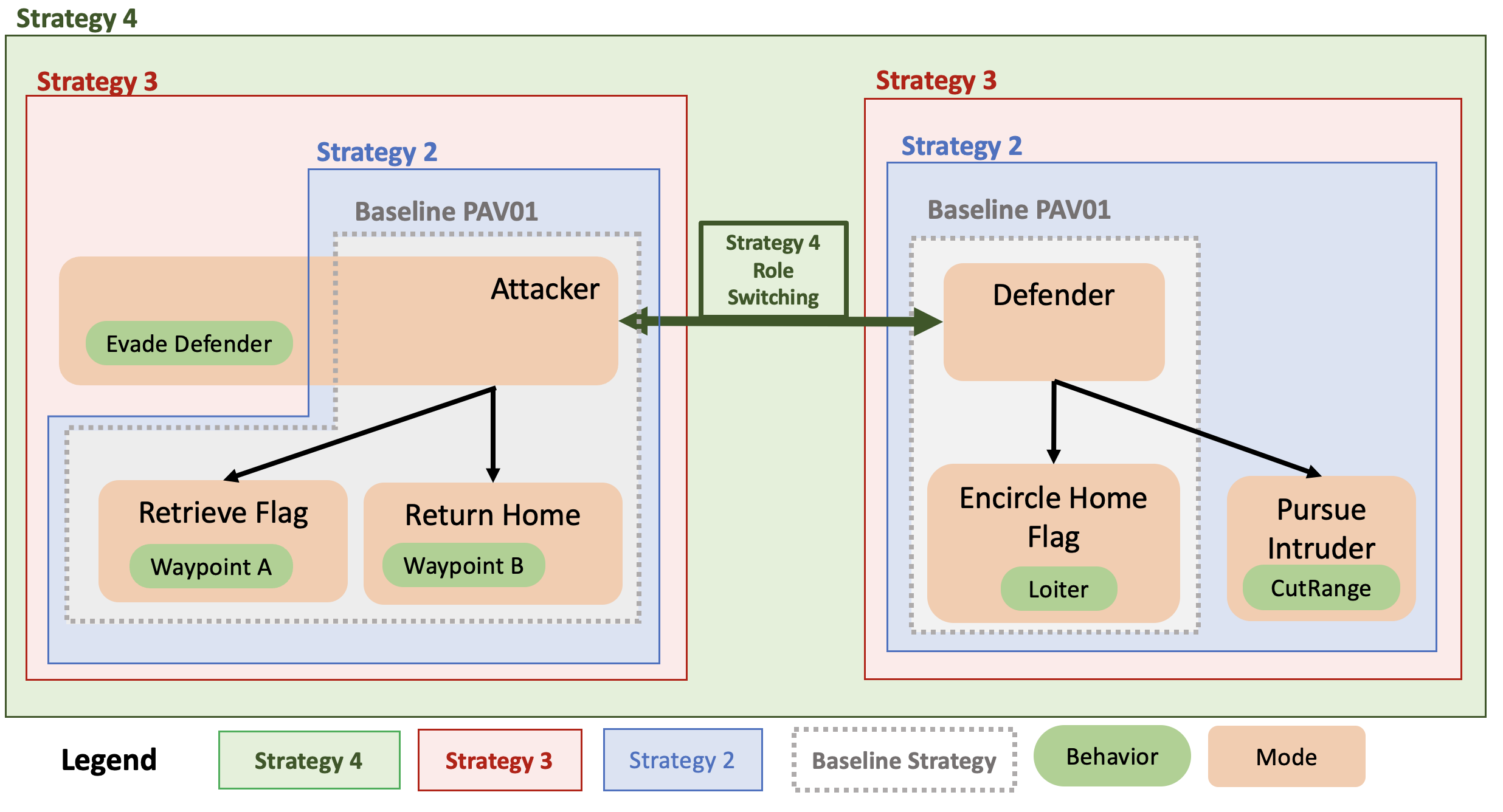}
    \caption{Mode tree for Behavior-Based Strategies. The OpRegion and AvoidCollision behaviors are not shown, but were always active. }
    \label{fig:bhv_strategies}
\end{figure}

\textbf{Strategy 2:} 
The attacker for this strategy is the same as for the baseline PAV01 strategy, \textbf{Easy Attacker}. The defender is assigned \textbf{Medium Defender}, it circles its flag to defend it from any enemies. If an enemy enters the defender's zone, the defender will chase after the enemy until it is tagged. The mode tree for this strategy can be seen in Figure \ref{fig:bhv_strategies}. This simple strategy was then expanded to implement more intelligent behavior. 

\textbf{Strategy 3:} The attacker was upgraded to be assigned \textbf{Medium Attacker}, to avoid any defenders in an effort to stay out of their tag range while also trying to reach the opposing flag. This trade-off was managed using multi-objective optimization in pHelmIvP. This upgrade demonstrated more intelligent behavior, and improved performance. See Figure \ref{fig:bhv_strategies} for the mode tree. After this, another upgrade was made to the behavior-based approach.

\textbf{Strategy 4:} By default, attacking agents were assigned the \textbf{Medium Attacker} and defending agents were assigned \textbf{Medium Defender}. Another upgrade to the approach was to use rule-based mode switching for both USVs on the team. The USVs would start and stay in attack mode until an enemy entered their zone. Once an intruder was detected, both USVs returned to tag the opposing agent before going back on the offensive. This strategy was designed to be an alternative and more aggressive strategy, allowing higher scores. It removed the mode in which a defender waited by encircling its own flag in an effort to better allocate the vehicle's time. However, the main advantage is that this strategy is agnostic to the number of vehicles, making it easily expandable. The mode tree may be seen in Figure \ref{fig:bhv_strategies}.

\subsubsection{Rule-based strategy Using Opponent Observation and Classification}

The conditional logic autonomy feature selects among these predetermined behaviors based on specific field conditions, which include:
\begin{itemize}
     \item Possession of the flag by a USV
     \item Positions of USVs
     \item Heading directions of USVs
     \item Tagging statuses of USVs
     \item Locations of the flags
\end{itemize}
Originally designed for $1$-v-$1$ engagements, this feature was adapted for $2$-v-$2$ scenarios by assigning each USV a distinct opposing USV. Despite this adjustment, each USV remains oblivious to its teammate and the teammate's opponent, reacting solely to its designated adversary. The operational logic unfolds as follows:
\begin{enumerate}
    \item An assessment is conducted to determine if the opposing USV adopts an offensive strategy.
    \begin{itemize}
        \item The USV adopts a defensive stance to observe whether the opponent crosses the midfield line, indicating an offensive posture, before proceeding to the next evaluation.
        \item In the absence of midfield crossing within approximately two minutes, the opponent is classified as non-offensive.
    \end{itemize}
    \item A subsequent assessment ascertains the aggressiveness of the opponent.
    \begin{itemize}
        \item An offensive opponent prompts the USV to continue its defensive stance. If the opponent directly approaches the flag, disregarding defensive maneuvers, it is deemed aggressive. Conversely, if the opponent attempts to circumvent the defending USV, it is classified as non-aggressive.
        \item If the opponent is defensive, the USV crosses midfield and pauses briefly. An opponent's pursuit categorizes them as aggressive; lack of pursuit signifies a non-aggressive disposition.
    \end{itemize}
    \item Strategy execution based on opponent classification:
        \begin{itemize}
            \item \textbf{Offensive - Aggressive:}
            The USV tags the opponent, exploiting their inattentiveness to defensive presence. Post-tagging, the USV advances towards the opponent's flag, timing its capture attempt based on the opponent-flag distance. Combination of \textbf{Medium Attacker} and \textbf{Medium Defender}.
            \item \textbf{Offensive - Non-Aggressive:}
            The USV maneuvers to force an evasive opponent out of bounds, reducing their speed due to safety protocols. This slowdown allows the USV to capture the opponent's flag. A combination of \textbf{Easy Attacker} and \textbf{Medium Defender}.
            \item \textbf{Non-Offensive - Aggressive:}
            This scenario presents a challenge. The USV probes the defense by entering and then retreating from the opponent's territory, aiming to lure the opponent away from their flag for a strategic advantage. A combination of \textbf{Medium Attacker} and \textbf{Medium Defender}.
            \item \textbf{Non-Offensive - Non-Aggressive:}
            The USV invades the opponent's territory, awaiting an opportune moment to capture the flag when sufficient distance between the opponent and their flag is achieved. \textbf{Easy Attacker} and \textbf{Medium Defender}.
        \end{itemize}
\end{enumerate}

\subsection{Pyquaticus Environment}

Pyquaticus \cite{pyquaticus} is a lightweight gymnasium environment written in Python intended for training RL agents to play aquatic CTF in simulation. The Pyquaticus environment has a few key aspects: (1) matching vehicle dynamics, action spaces, observation spaces, and game rules with the MOOS-IvP Aquaticus Test-bed, (2) direct integration with RLLib \cite{rllib_cite}, PettingZoo \cite{pettingzoo}, and Stable Baselines \cite{stable-baselines}, and (3) a parallelized structure, allowing it to run many instances simultaneously. The key to the Pyquaticus environment is that it can be used within the MOOS-IvP framework as a behavior to allow velocity- and behavior-based control of marine vehicles using previously learned policies. This greatly reduces the overhead of deploying a machine learning-based control policy on the Aquaticus setup and allows Pyquaticus agents to also run in pHelmIvP simulations. 

\begin{figure}
    \centering
    \includegraphics[width=1\linewidth]{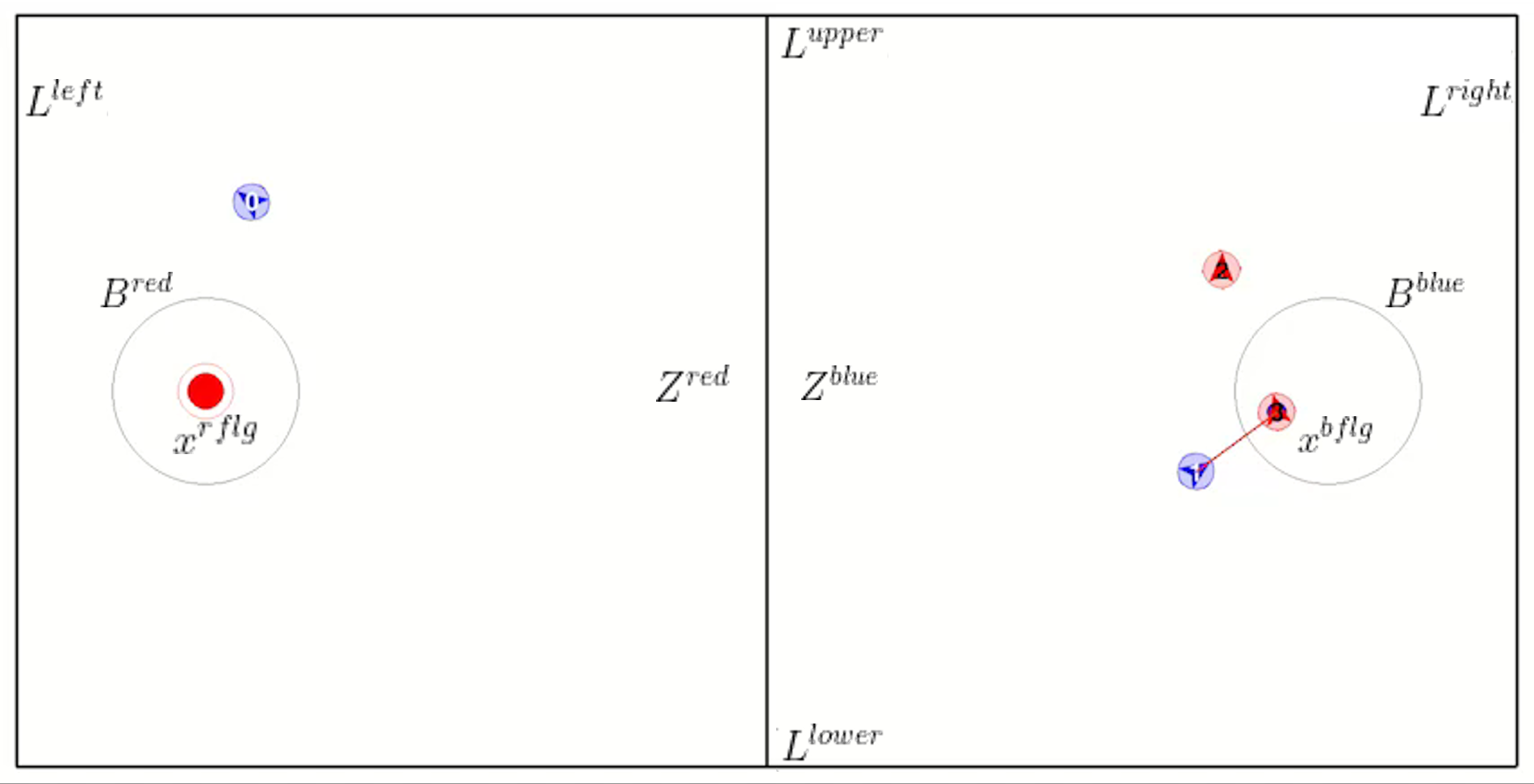}
    \caption{Pyquaticus rendering in pygame with two agents on each team. The red team has captured the blue team's flag, and the blue team is near the red team's exposed flag.}
    \label{fig:Pyquaticusl}
\end{figure}

\subsubsection{Formal Game Representation}
We extend the formal game representation of Aquaticus CTF~\cite{Kliem2023} from a $1$-v-$1$ attacker versus defender scenario to a $2$-v-$2$ scenario where players have to learn both attacking and defending roles. The environment of the Aquaticus CTF game is represented by a rectangular, obstacle-free playing area of width $W$ and depth $D$ bounded by a set of four straight lines, $\mathbf{L}^{bounds} = \{L^{lower}, L^{upper}, L^{left}, L^{right}\}$. The two team zones are represented by $Z^{blue}$ and $Z^{red}$, and, each team's respective base regions are represented by $B \subset Z$ and $B^{red} \subset Z^{red}$, respectively. Flag locations start within the respective team's bases and are denoted by $\mathbf{x^{bflg}}$ and $\mathbf{x^{rflg}}$. A game consists of two teams of two players each, denoted by $\mathbf{T^{blue}}$ and $\mathbf{T^{red}}$, respectively. Each team has two players, e.g. $\mathbf{x_i^{blue}}$ where $i \in \{0, 1\}$ is the agent-id. For the remainder of the paper and for the sake of legibility, we have dropped the team color (blue or red) in the notations and definitions, assuming that it is understood from context. 

An agent's state set is given by $S_i = \{(\mathbf{x_i}, \theta_i)\}$, where $\mathbf{x_i} = (x_i, y_i) \in \mathbb{Z}^2$ specifies its location in the playing field. $\theta_i \in \Theta = \{\theta_1, \theta_2, \ldots, \theta_K\}$ is a discretization of the angular bearing, $[0, 360]$ into $K$ disjoint segments. The state space of the game is represented as the joint state sets, $S = \times_{i = \{0, 1\}}S_i \times S_i^{red}$. A subset of states, called game event states, denoted by $S^{EV} = \{S^{tag},  S^{grb}, S^{cap}\}$, correspond to the point-scoring game events. The action set of an agent is specified as $A_i = \{(\nu, \theta)\}$, where $\nu \in \{0 \cup \Re^+\}$ and $\theta \in \Theta$ are an agent's velocity and heading, respectively. The action space of the game, $A$, is given by the joint actions of agents on both team's. The forward dynamics model or transition function ${\mathbb T}: S \times A \times S \rightarrow [0,1]$ gives a probability distribution for the next game state given the current game state and current actions taken by the players. An agent reaching any of the game event states receives a reward, $R: S^{EV} \times A_i \rightarrow \Re$, as seen in Table~\ref{tab:ctf_game_outcomes}. Rewards at all other states are $0$. Finally, agent $i$'s policy $\pi_i: S \rightarrow \Pi(A_i)$, is a mapping that takes the current state of the game and returns a probability distribution over the action set of agent $i$.

\subsubsection{Pyquaticus Reward Functions}
Each agent starts from an initial state $s_{i, 0}$ located inside its home base. At time-step $t$, $(s_{i, t}, a_{i, t})$, and $r_{i, t}$ denote the agent's state, action and reward received. The problem facing agent $i$ is to find an optimal policy $\pi_i^* = \displaystyle \max_\pi {\mathbb E}[\sum_{t=0}^{H} \gamma^t R(s_t, a_{i, t})]$, where $a_{i, t}) = max_{a \in A_i} \pi(a|s_t)$, $s_{t+1} = max_{s \in S_i} T(s_t, a_{i, t}, s)$ and $H$ is the horizon or the number of time steps corresponding to the duration of the game. We use multiagent reinforcement learning algorithms to solve each agent's policy maximization problem, as described in the next section.

The default reward function in pyquaticus is given as, 
\begin{equation}
    R(s_t, a_{i, t})= S^{grb}_{opp} + S^{cap}_{opp} + S^{grb}_{own} + S^{cap}_{own} + S^{tag}_{own} + S^{tag}_{opp} + L^{bounds}
\end{equation}

The default values for the reward function are in Table \ref{tab:ctf_game_outcomes}. The Deep Reinforcement Learning approaches used modified versions of the default reward function and event reward values.

\begin{table}[h!]
    \centering
    \caption{Baseline Sparse Rewards for the Aquaticus CTF game.}
    \begin{tabular}{|c|c|c|c|}
        \hline
        {\bf Event} & {\bf Description} & {\bf Team } & {\bf Opp. Team}\\
        {\bf Name} & {} & {\bf Reward} & {\bf Reward} \\
        \hline
        {Player} & {Agent gets within $10$ m} & {$100$} &  {$-100$} \\
        {Tag} & {of opposing agent while} & {} & \\
        {(No Flag)} & {in their team's zone } & {} & \\
        \hline
        {Player} & {Agent gets within $10$ m} & {$50$} &  {$-100$} \\
        {Tag} & {of opposing agent while} & {} & \\
        {(With Flag)} & {in their team's zone } & {} & \\
        \hline
        {Flag} & {Agent grabs opponent's } & {$50$} & {$-50$}\\
        {Grab} & {flag from opponent's base} & {} & {}\\
        \hline
        {Flag} & {Agent returns to their team's side} & {$100$} & {$-100$}\\
        {Capture} & {base with opponent's flag} & {} & {}\\
        \hline
        {Out of Bounds} & {Agent drives out of the game field} & {-100} & {0} \\
        \hline
        
    \end{tabular}
    \label{tab:ctf_game_outcomes}
\end{table}

\subsubsection{Deep Reinforcement Learning using Pyquaticus}
Multiple policies were trained using various reinforcement learning-based approaches. Fundamentally, all of the approaches employed fell into three categories: (1) direct on-policy RL using Proximal Policy Optimization (PPO) \cite{schulman2017proximal} or (2) direct off-policy RL using Twin Delayed DDPG (TD3) \cite{fujimoto2018addressing} with velocity-based action spaces, and (3) options-based hierarchical RL \cite{sutton1999between} over RL-trained versions of built-in Aquaticus behaviors. 

The PPO and TD3 implementations relied heavily on large volumes of training data but often struggled to learn safety constraints (e.g., the environment boundary) and took a very long time to develop any reasonable behaviors. These policies often acted extremely defensively and prioritized defense of their own flag over taking an opponent's flag. This is likely because they were rarely able to acquire an opponent's flag, and even more rarely able to capture it. The reward signals were also not dense enough to motivate policies to take any risk, and the credit assignment problem was difficult to solve at the low level. This is because there were relatively long action traces before an attack-incentivizing reward signal was discovered. 

The options-based framework used a modified options-critic deep Q-learning approach to train agents to select a behavior from a set of policy options at each timestep during the game. These policy options (pickup opponent flag, guard own flag, tag opponent, avoid opponents, retreat, shield teammate from opponent tag) were pre-trained with Deep Double Q-learning \cite{van2016deep}. Choosing between behavior primitives, rather than low-level control inputs, allows the RL process to more easily attribute specific choices to specific successful (or unsuccessful) outcomes.

\omitit{
\begin{table}[h!]
    \centering
    \caption{Sparse Reward Function}
    \begin{tabular}{|c|c|}
        \hline
        {\bf Reward Name} & {\bf Value} \\
        \hline
        {Opponent Tag} & {$-100$} \\
        \hline
        {Opponent Flag Grab} & {$50$} \\
        \hline
        {Opponent Flag Capture} & {$200$}\\
        \hline
        {Player Tag} & {$100$} \\
        \hline
        {Team Flag Grab} & {$-50$} \\
        \hline
        {Team Flag Capture} & {$-200$} \\
        \hline
    \end{tabular}
    \label{tab:sparse_reward_function}
\end{table}}

\section{Results and Implications from October 2023 Experiments}
\subsection{Overview}The USV CTF competitions resulted in a total of 22 games played with a total of over 3.5 hours of gameplay. Six teams from backgrounds in academia and defense participated in the event, which was facilitated by the United States Military Academy's Robotics Research Center with assistance from the Massachusetts Institute of Technology's (MIT) Marine Autonomy Lab (PavLab). Participants such as MIT Lincoln Laboratory, U. S. Naval Research Laboratory, the UK's Defence Science and Technology Laboratory (DSTL) and Australia's Defence Science and Technology Group (DSTG), utilized the Pyquaticus library for implementing Deep RL within the Aquaticus test-bed, while MIT PavLab utilized features of the pHelmIvP for hierarchy behavior-based autonomy. 
The following table shows a breakdown of the AI methods opposed, their respective teams, and overall competition results. 




\begin{table}[h]
    \caption{CTF metrics from Australia USV competitions. Green and red cells indicate highest and lowest average totals.}
    \centering
    \resizebox{\columnwidth}{!}{%
    \begin{tabular}{ccccc}\toprule
         AI Method & Default(Pav01) & Rule-based & Pyquaticus\\
         \midrule
         Games Participated &  9 & 15 & 20 \\
         Mean GRABS per Game & 3 & 4\cellcolor{green!15} & \cellcolor{red!15}$<$1 \\
         Mean CAPTURES per Game & 1 &\cellcolor{green!15} 2 & \cellcolor{red!15}$<$1 \\
         Mean TAGS per Game &  2& 3 \cellcolor{green!15}&3 \cellcolor{green!15}\\
         \bottomrule
         Mean Score per Game &  5& 8 & $<$1\\ \bottomrule
    \end{tabular}
   }
    \label{tab:Competition_results}
\end{table}

\subsection{Initial Takeaways}
Table \ref{tab:Competition_results} shows that algorithms using pHelmIVP behavior-based autonomy had a significant advantage over both the default Pav01 entry and the Deep RL autonomy methods that used Pyquaticus. Interestingly, Pyquaticus Deep RL models tended to perform in an extremely defensive manner, either hugging side borders or loitering near their own flag to prevent it from being taken. While this method did cause the mean number of captures to significantly decrease for both pHelmIvP and default opponents, flag grabs still resulted in a much higher average score for the hierarchy behavior-based autonomy solutions. 

An investigation into which specific pHelmIvP based algorithms were the most successful revealed that rule-based cooperation (changing roles) among agents resulted in a robust game strategy that proved effective versus both the default entry and the Pyquaticus Deep RL opponents. Unlike Deep RL where the agents are trained against themselves, pHelmIvP with rule-based cooperation only uses knowledge of the opponents' activities to determine actions. If no opponents were within friendly bounds, both agents would switch into attack mode. The agents would return to defense when an opposed agent attempted to grab their flag. This hierarchy policy resulted in a teaming-like effect, where the agents would take immediate advantage of tagging out a defending opponent by both subsequently attacking the flag now defended by a lone agent. 
\begin{figure}[h]
    \centering
    \includegraphics[width=.75\linewidth]{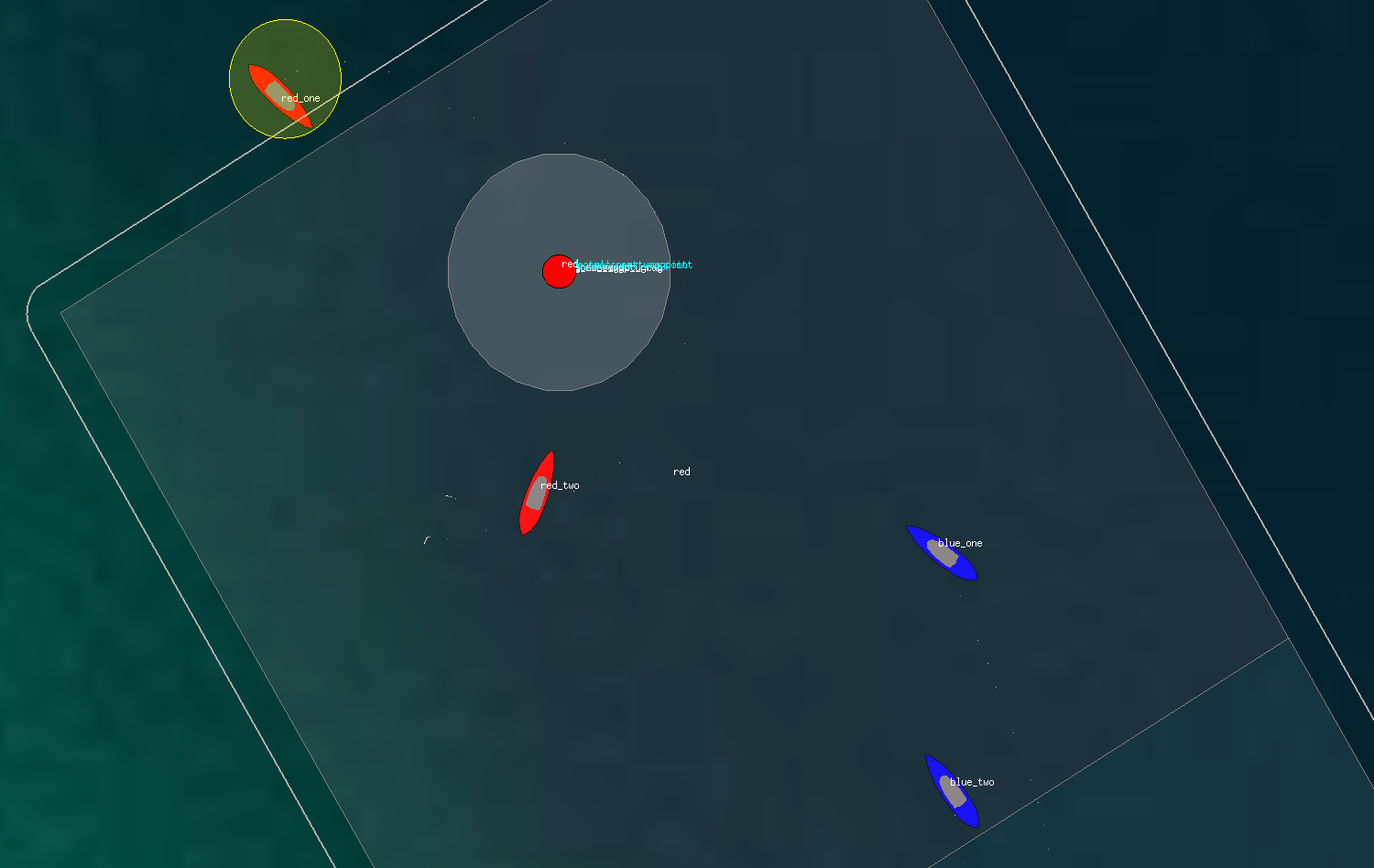}
    \caption{Blue team: Helm-IvP Behavior-Based Strategy 4, attacking a Pyquaticus-based opponent with two attackers.}
    \label{fig:BLUETEAM}
\end{figure}



Additionally, since the pHelmIvP behavior-based optimizer ability to solve multiple behaviors simultaneously enables grouping of concurrent behaviors into a hierarchy of modes, the control of switching between states in a given mode and the switching between modes in a given hierarchy of states can be explicitly controlled by rules which can be learned by agents playing the CTF game.

Pyquaticus policies trained with PPO or TD3 were deployed against one another often guarded their respective flags and rarely engaged with the opponent. This is interpreted to be a result of the long action traces before an attack-incentivizing reward signal was discovered. Unsurprisingly, better performance and faster training times were shown when a curriculum learning scheme was implemented; however, all of these policies remained greedy and conservative in their game strategies. 

Because this options-based approach did not require the agent to learn vehicle dynamics, it was able to more quickly learn policies that exhibited meaningful behavior. These policies were slightly less conservative than the PPO- or TD3-learned control policies. This stemmed from the benefit of being much more stable and predictable at the control level because they are selecting between predetermined behavior primitives. However, the options framework would often over-prioritize defending the flag, or doing nothing as opposed to navigating into the opponents area. Therefore, finding a set of reward signals that emphasize more aggressive policies is the focus of near-term future work.

\subsection{Challenges and Future Works}
The current findings of the Aquaticus competitions demonstrate a significant support for hierarchical behaviors. However, the authors argue that additional work is necessary to address the challenges faced during the competition and explore various topics in Deep RL before a definitive conclusion can be drawn. Logistics-related changes to the testing location caused several difficulties within the Pyquaticus model configuration files, resulting in a decreased number of CTF games played and a lack of proper integration with AvoidCollision pHelmIvP behaviors. Furthermore, external elements like strong winds and rough sea conditions led several Pyquaticus USVs to be pushed beyond the boundaries because of their tendency to stick to the edges of the playing field during the competition. Further integration of Pyquaticus with MOOS-IvP in terms of configuration and control schema of the USVs will allow for more competitive CTF games in future studies. 

In addition to further integration work of Deep RL in the Aquaticus test bed, a number of novel research concepts can be explored in future Pyquaticus studies over a number of research topics such as reward shaping, training optimization, and further investigation of "sim-to-real" inconsistencies. Research will continue on efforts at extending rule-based cooperation among agents to react to safety and security violations in accordance with human experts intent/rules for executing safety and security process behaviors by exploiting the pHelmIvP behavior-based optimizer ability to solve multiple behaviors simultaneously to group concurrent safety and security behaviors into a hierarchy of CTF behavior modes.

\section{Conclusion}  
This work has presented a novel approach to evaluating cooperative teaming autonomy via the Aquaticus test-bed. The implemented CTF competition presents a real-time strategy game in which agents benefit from teaming behaviors and resiliency to opponents from different autonomy backgrounds. As robotic vehicles working in cooperative teams become more prevalent in civilian and military contexts, this research will help identify which methods of multi-robot autonomy perform best and why. Although the results in this paper were mostly discussed on a qualitative level due to ongoing engineering efforts, this work will serve as a stepping stone for future Aquaticus test-bed experiments. Initial takeaways imply that rule-based cooperative autonomy works in a robust manner due to MOOS-IvP's hierarchical behavior configurations. Rule-based behaviors are based solely on in-game knowledge to perform, as opposed to other methods that specifically train against another opponent. Deep RL methods such as PyQuaticus have very promising qualities that have resulted in defensive methods. These emergent behaviors are predicted to change with more aggressive reward shaping and the restructuring of specific configuration challenges related to competition. Future works in the Aquaticus project are planned with an emphasis on quantitative metrics to discuss key differences of behavior-based and Deep RL models.

        



\bibliographystyle{IEEEtran}
\bibliography{IEEEabrv,export}


\end{document}